%% file: acl_latex.tex
\pdfoutput=1

\documentclass[11pt]{article}

\usepackage[final]{acl}

\usepackage{times}
\usepackage{latexsym}
\usepackage{amsmath}
\usepackage{amsfonts}
\usepackage{siunitx}
\usepackage{tabularx}
\usepackage{makecell}
\usepackage[table]{xcolor}
\usepackage[T1]{fontenc}

\usepackage[utf8]{inputenc}

\usepackage{microtype}

\usepackage{inconsolata}

\usepackage{graphicx}
\usepackage{algorithm}
\usepackage{algpseudocode}
\usepackage{booktabs}
\usepackage{multirow}
\usepackage{tcolorbox} 
\newtcolorbox{mybox}{fontupper=\footnotesize}

\usepackage[normalem]{ulem}

\usepackage{tabularx}
\usepackage{booktabs}
\usepackage{multirow}

\usepackage{makecell} 

\usepackage{tikz}
\usetikzlibrary{shapes, arrows, positioning, backgrounds}
\usepackage{adjustbox}

\usepackage{amsmath,amssymb}   
\usepackage{enumitem}          

\title{V-VAE: A Variational Auto Encoding Framework Towards \\Fine-Grained Control over Human-Like Chat}

\author{
  Qi Lin$^{1}$, Weikai Xu$^{1}$, Lisi Chen$^{1*}$ and Bin Dai$^{2*}$\\
  \normalsize $^1$University of Electronic Science and Technology of China \\
  \normalsize $^2$Qingyao Intelligence \\
  \normalsize linqi@std.uestc.edu.cn, \normalsize xuwk266@gmail.com \\
  \normalsize lchen012@e.ntu.edu.sg, \normalsize daibin@jinyao.tech \\
}

\begin{document}
\maketitle
\renewcommand{\thefootnote}{\fnsymbol{footnote}}
    \footnotetext[1]{Corresponding authors: Lisi Chen and Bin Dai.}
\begin{abstract}
With the continued proliferation of Large Language Model (LLM) based chatbots, there is a growing demand for generating responses that are not only linguistically fluent but also consistently aligned with persona-specific traits in conversations.
However, existing role-play and persona-based chat approaches rely heavily on static role descriptions, coarse-grained signal space, and low-quality synthetic data, which fail to capture dynamic fine-grained details in human-like chat. 
Human-like chat requires modeling subtle latent traits, such as emotional tone, situational awareness, and evolving personality, which are difficult to predefine and cannot be easily learned from synthetic or distillation-based data.
To address these limitations, we propose a \textbf{\underline{V}}erbal \textbf{\underline{V}}ariational \textbf{\underline{A}}uto-\textbf{\underline{E}}ncoding (\textbf{V-VAE}) framework, containing a variational auto-encoding module and fine-grained control space which dynamically adapts dialogue behaviour based on fine-grained, interpretable latent variables across talking style, interaction patterns, and personal attributes. 
We also construct a high-quality dataset, HumanChatData, and benchmark HumanChatBench to address the scarcity of high-quality data in the human-like domain. 
Experiments show that LLMs based on V-VAE consistently outperform standard baselines on HumanChatBench and DialogBench, which further demonstrates the effectiveness of V-VAE and  HumanChatData.
\end{abstract}

\input{sec/sec_intro}

\input{sec/sec_related_work}

\input{sec/sec_method}

\input{sec/sec_exp}

\input{sec/sec_conclu}

\newpage
\section*{Acknowledgment}
This work was supported by the National Key R\&D Program
of China (No. 2023YFC3305600).
\section*{Limitations}
Despite the effectiveness of our approach, we acknowledge two limitations:
First, although latent space representations have demonstrated empirical effectiveness, they still lack a well-defined theoretical framework or principled criteria for organizing and summarizing fine-grained attributes. Moreover, such subtle latent features often challenge human annotators to perceive or label consistently, limiting the efficiency of annotation and the interpretability of human-supervised signals.
Second, human annotations inherently carry subjective preferences and inductive biases, especially in tasks involving dialogue quality or persona alignment. Such biases may reduce the generalizability of the trained models, particularly when deployed in domains with divergent user expectations or cultural norms.
\section*{Ethics Statement} 
We have rigorously refined our dataset to remove any elements that could compromise personal privacy, thereby guaranteeing the highest level of protection for individual data. All data annotations were completed by crowdsourced volunteers, to whom we paid \$0.5 per step as compensation and provided the necessary training. The human evaluation of our work was carried out through a meticulously randomized selection of IT professionals. This process ensured a gender-balanced and educationally diverse panel, reflecting a wide spectrum of perspectives and expertise.
\bibliography{custom}

\input{sec/sec_appendix}

\end{document}

%% file: sec/sec_intro.tex
\section{Introduction}


LLM-based chatbots are becoming increasingly popular and intelligent. Their applications are ubiquitous in a variety of specialized domains, such as online education~\citep{chang2025review}, customer service~\citep{park2024align}, and digital human~\citep{suo2025digital}. These chatbots are expected to follow specific roles during human-chatbot interactions, requiring them to exhibit personalized attributes and characteristics associated with those roles~\citep{tamoyan2024llm}.
For this purpose, existing studies aim to improve the control over persona-related attributes by employing predefined role documents through in-context learning~\citep{chen2023places}, limiting the length of individual responses in the decoding phase and the number of dialogue rounds~\citep{zhao2024long}, and modifying interaction styles in the training corpus~\citep{shen2024rethinking}. 
With chatbots now capable of assuming designated roles, recent efforts~\citep{ccalik2025enhancing} aim to further enhance their ability to exhibit human-like behavior in LLM-driven conversations.
\begin{figure} [t]
\setlength{\abovecaptionskip}{0.3cm}
\setlength{\belowcaptionskip}{-0.3cm}
    \centering
    \includegraphics[width=8cm]{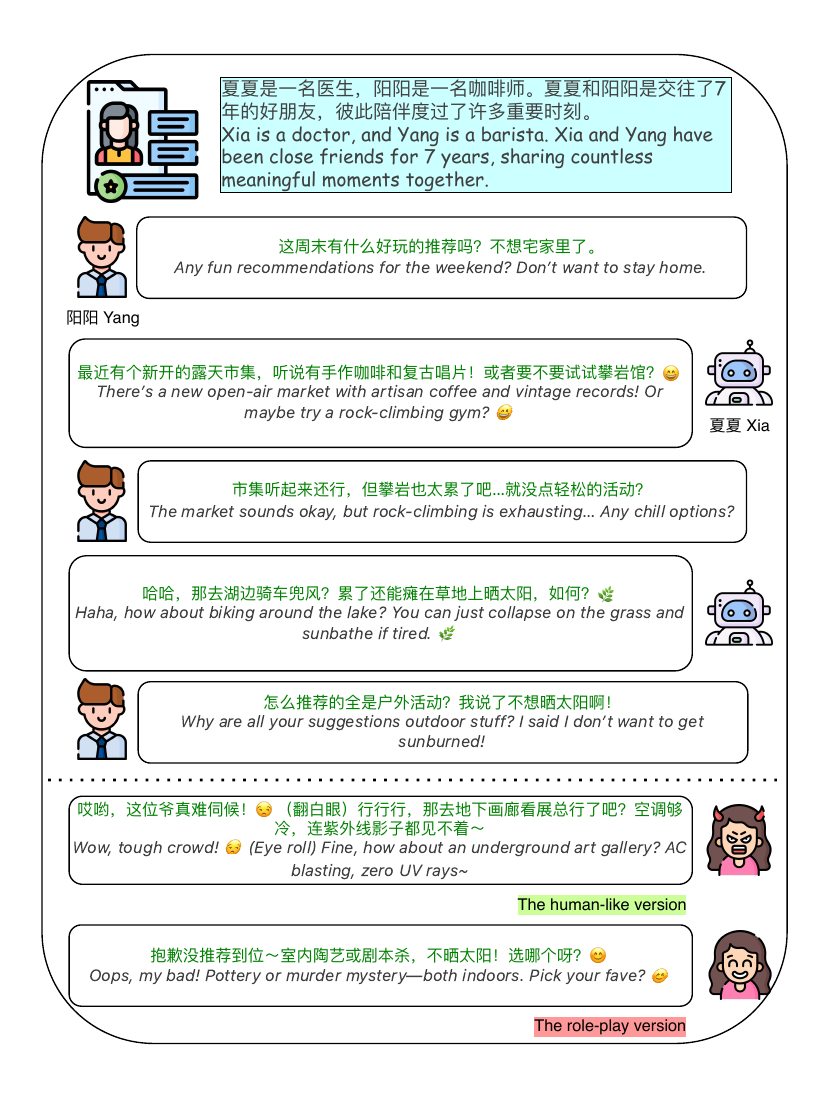}
    \caption{The difference between role-play chat and human-like chat responses.}
    \label{fig:history}
\end{figure}

However, as shown in Figure~\ref{fig:history}, human-like chat imposes more stringent requirements compared to role-play chat~\citep{tu2024charactereval} and persona-based chat~\citep{yamashita2023realpersonachat}, as it needs to capture more authentic human interaction histories and exhibit a broader range of conversational attributes.
Specifically, human-like chat modeling poses the following three challenges:
(1) Unlike static and predefined role descriptions, human-like chat unfolds dynamically, with the chatbot's tone, lexical choices, and interaction patterns evolving for the communication process. 
(2) While role-play chat relies on scripted constraints and persona-based conversations use explicit labels (e.g., age, occupation) to define character traits explicitly, human-like chat involves complex and latent signals such as emotional tendencies, situational awareness, and evolving personality traits. These aspects are difficult to model from initialization and even harder to evaluate quantitatively.
(3) Role-play chat exhibits a weaker dependency on such latent signals, as its generation is based on human-AI interaction for data synthesis~\citep{kim2024understanding} or direct teacher chatbot distillation~\citep{hu2025efficient}. 
In contrast, the higher quality demands of human-like chat data, especially its fidelity to real human behavior, make it impractical to build using automated pipelines or human-machine dialogue platforms.

To address the above challenges, we propose a Verbal Variational Auto-Encoding (V-VAE) framework consisting of the following three key components:
(1) \textbf{Variational Auto-Encoding Mechanism}: To overcome the rigidity of static, predefined role specifications, we develop a variational mechanism that dynamically encodes and updates role-relevant information as the chat progresses. This allows the chatbot to adapt its interaction patterns in response to evolving conversational context. 
(2) \textbf{Fine-Grained Latent Space}: We decompose the dialogue control space into three orthogonal dimensions: talking style, interaction patterns, and personal attributes. This structured latent space enables more precise and interpretable control over chatbot behaviour. Further, to evaluate the human-likeness of generated responses, we build three new metrics named Catchphrase Presence (\textit{CP}), Emoji Consistency (\textit{EC}), and Hobby Mentioning (\textit{HM}). 
(3) \textbf{Human-Like Chat Dataset}: We construct a new human-like chat dataset named HumanChatData, and an evaluation benchmark HumanChatBench, through comprehensive human annotation. This dataset addresses the scarcity of high-quality training data for human-like conversational modeling.
Experiments on HumanChatBench and public human-like chat DialogBench show that V-VAE based LLMs outperform backbones consistently and even surpass close-source LLMs.
In particular, Qwen-VVAE achieves an average improvement of 7.2\% over Qwen-7B on the human-likeness metrics defined by DialogBench.

In summary, our contributions are threefold.

$\bullet$ We propose Verbal Variational Auto-Encoding (V-VAE), a dynamic framework that analyzes and adjusts chat behaviours automatically, breaking the limitations of conventional role-based dialogue ingrained patterns.

$\bullet$ We develop a Fine-Grained Latent Space, a more expressive and structured template for controlling human-like dialogue styles. This design enables more accurate modeling of subtle and implicit features in multi-turn chat.

$\bullet$ We construct HumanChatData, a high-quality human-like dialogue dataset, and propose HumanChatBench, an accompanying evaluation benchmark. This effort helps bridge the gap in high-quality training data for human-like chat.



%% file: sec/sec_related_work.tex
\section{Related Work}

\subsection{Human-like Chat} 
A variety of research has been dedicated to enhancing the human-like qualities of large language model (LLM) responses~\citep{li2023reinforcement,ccalik2025enhancing}. 
Techniques such as Reinforcement Learning from Human Feedback (RLHF) have significantly refined model outputs by aligning them with user preferences and expectations~\citep{cuayahuitl2019deep,jaques2020human}.
One prominent model, DialoGPT~\citep{zhang2019dialogpt}, leverages extensive Reddit data to produce responses that closely resemble human conversation. 
Similarly, Meena, a multi-turn chatbot, has been optimized to achieve high dialogue coherence through metrics like Sensibleness and Specificity Average (SSA)~\citep{durmus2023towards,adiwardana2020towards}. 
For benchmarks, several datasets were released to evaluate the human-likeness of chatbot in the field of human-robot interaction~\citep{kahn2007human,duan2024hlb,ying2025benchmarking,ou2023dialogbench}.
Our work emphasizes the nuanced integration of real-time emotional states, relational dynamics, and interactional context between dialogue participants, dimensions often overlooked in conventional emotion-aware systems, which have been ignored in previous work.

\begin{figure*} [ht]
    \centering
    \includegraphics[width=1\textwidth]
    {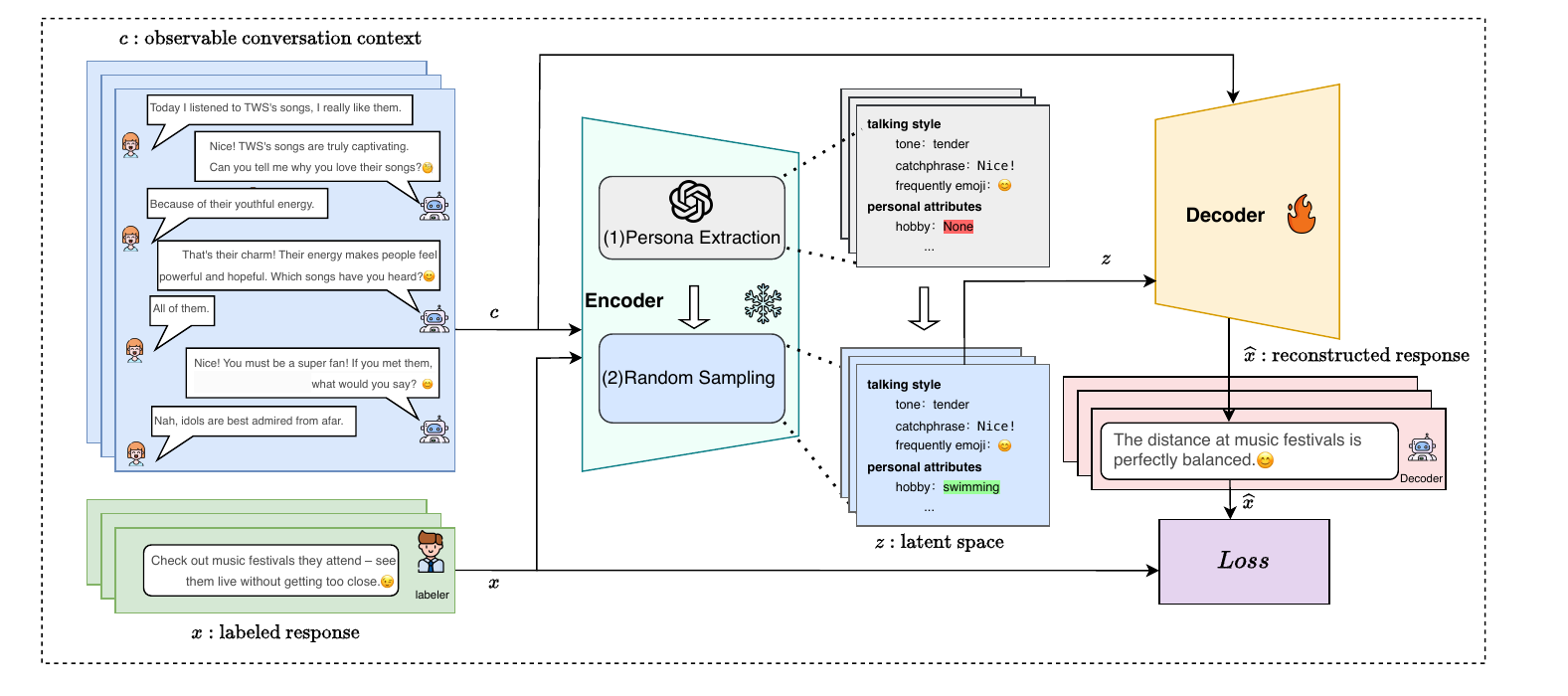}
    \caption{An overview of the V-VAE framework, which adopts an encoder–decoder architecture for latent-variable conditional generation. The encoder comprises two components: persona extraction and prior-based sampling (e.g., when the AI's hobby is unobserved in the conversation context, it is sampled from the prior distribution). The decoder reconstructs responses conditioned on both the extracted persona and the dialogue context, and is trained by minimizing the loss between the reconstructed response and the ground-truth target.
}
    \label{fig:main}
\end{figure*}


\subsection{Human-likeness}
Prior research~\citep{danesi2017language, le2017evolution} has established foundational sociolinguistic influences spanning identity construction, digital communication patterns, and multimodal semiotics like emoji usage. While large language models (LLMs) derive training data predominantly from naturalistic texts, their inherent one-to-many deployment paradigm—wherein a single model serves heterogeneous user bases—engenders systematic objectivity in politeness conventions and generates perceptibly AI-like linguistic outputs. Consequently, significant scholarly investigations have quantified sociolinguistic divergences between LLM and human communication\citep{schneider2024sociolinguist}. These variations manifest across three core dimensions: Talking style, encompassing emoji interpretation and deployment dynamics\citep{zheng2025irony} and measurable disparities in grammatical and rhetorical composition~\citep{schneider2024sociolinguist, reinhart2025llms}; Interaction patterns, characterized by relational alignment failures during longitudinal interactions\citep{altenburger2024examining}; Personal attributes, evidenced in absent human-like psycholinguistic properties in extended discourse\citep{seals2023long}. Our feature space design is therefore grounded in these empirically validated dimensions of cross-modal variation, systematically addressing each categorical divergence through targeted representation learning.





%% file: sec/sec_method.tex
\section{Method}
\label{sec:method}
\subsection{Task Formulation}
We formulate human-like chat generation as a latent-variable conditional generation task. Given an observable conversation context $c$, that a labeler can refer to when writing the response $x$, the goal is to maximize the likelihood of $x$ given the condition $c$. Besides the context $c$, there are also some unobservable variables that controls the generation of $x$, including but not limited to tone, frequently-emoji and personal hobbies. We denote it as the latent variable $\mathbf{z}$, which is a sample from the latent space $\mathcal{Z}$.
We define the generation model as $p_\theta(x \mid c, z)$, and aim to maximize the likelihood of responses in a dataset $\{(x_i, c_i)\}_{i=1}^N$, where $N$ is the size of the dataset. 
\begin{equation}
    -\log p_\theta(x|c) = -\log \sum_{z\in\mathcal{Z}} p_\theta(x|c,z)\cdot p_\lambda(z),
    \label{eqn:gt_obj}
\end{equation}
Here, $p_\lambda(z)$ denotes the prior distribution over the latent variable $z$ parameterized by $\lambda$.

\subsection{Variational Auto-Encoding Mechanism}
However, directly optimizing Equation(~\ref{eqn:gt_obj}) is intractable. To address this, we propose the V-VAE framework, which models response generation using a latent-variable encoder–decoder architecture, as illustrated in Figure~\ref{fig:main}. The encoder integrates both explicit persona cues and sampled latent attributes, allowing for flexible representation of speaker characteristics even when certain traits are unobserved. The decoder then conditions on both the inferred persona and the conversational context to reconstruct the target response, and is trained via reconstruction loss.

To make the objective tractable, we introduce a variational posterior distribution $q_\phi(z \mid x, c)$, we assume that $z$ is independent of the observable context $c$ and derive a variational upper bound of (\ref{eqn:gt_obj}) for $-\log p_\theta(x \mid c)$:
\begin{align}
    \geq &\sum_{z\in\mathcal{Z}} -q_\phi(z \mid x, c) \cdot \log p_\theta(x \mid c, z) \notag \\
    & + \mathbb{KL}\left[q_\phi(z \mid x, c) \,\|\, p_\lambda(z)\right] \label{eqn:elbo}
\end{align}
Note that Equation(~\ref{eqn:elbo}) is valid for any $q_\phi(z|x,c)$. The details of the derivation can be found in the appendix. Specifically, we define 
\begin{equation}
\label{eq:posterior_condition3}
q_\phi(z|x,c) = \prod_{k=1}^K q_\phi(z_k|x,c)
\end{equation}
Where $z_i$ is one predefined aspect of the latent space. And we introduce a structured latent variable $\mathbf{z} = [z_1, \dots, z_K]$, which encodes unobservable yet influential factors such as talking style, interaction patterns, and personal attributes. The latent prior space $\mathcal{Z}$ in Equation(~\ref{eqn:elbo}), which defines the full set of possible values for $z$, is predefined as the Cartesian product of discrete subspaces:
\begin{equation}
    \mathcal{Z} = \prod_{k=1}^K \mathcal{Z}_k
\end{equation}
Where $|\mathcal{Z}_k|$ is the cardinality of the space $\mathcal{Z}_k$, $K$ is the number of latent aspects and $\mathcal{Z}_k$ is the $k$-th sub latent space. $\mathcal{Z}_k$ is a prior closed set, the cardinality of $\mathcal{Z}_k$ can differ for different $k$. For example, \emph{relationship} can serve as one latent dimension, where the corresponding subspace is defined as $\mathcal{Z}_{\text{relationship}} = \{\text{stranger}, \text{acquaintance}, \text{enemy}, \text{lover}, \text{enemy}, \dots\}$. 



We use an existing powerful LLM $\pi_\phi(\cdot)$ to define the variational posterior. Specifically, we write a proper prompt that takes both the observable context $c$ and the response $x$ as input and ask the LLM what the value of the latent aspect is. If the response dose contain any information about that aspect, the LLM, if powerful enough, can give the correct answer $\pi_\phi(x,c)$. If the information is not included, we can suggest the LLM to output $\pi_\phi(x,c)=\emptyset$. Thus, we can define the posterior as
\begin{equation}
q_\phi(z_k|x,c) =
    \begin{cases}
    1 & \text{if } z_k = \pi_\phi(x,c) \neq \emptyset \\
    0 & \text{if } z_k \neq \pi_\phi(x,c) \neq \emptyset \\
    p_\lambda(z_k) & \text{if } \pi_\phi(x,c)=\emptyset
    \end{cases}
\label{eqn:posterior}
\end{equation}

Since we are using a fixed encoder (with properly designed prompt), the KL divergence in (\ref{eqn:elbo}) is not related to the parameters $\theta$. Thus we can omit the KL divergence and the final objective then becomes
\begin{equation}
    \mathcal{L} = \mathbb{E}_{z\sim q_\phi(z|x,c)}[-\log p_\theta(x|c,z)].
    \label{eqn:final_objective}
\end{equation}





\subsection{Design of Latent Persona Space}

\begin{figure} [ht]
    \centering
    \includegraphics[width=8cm]{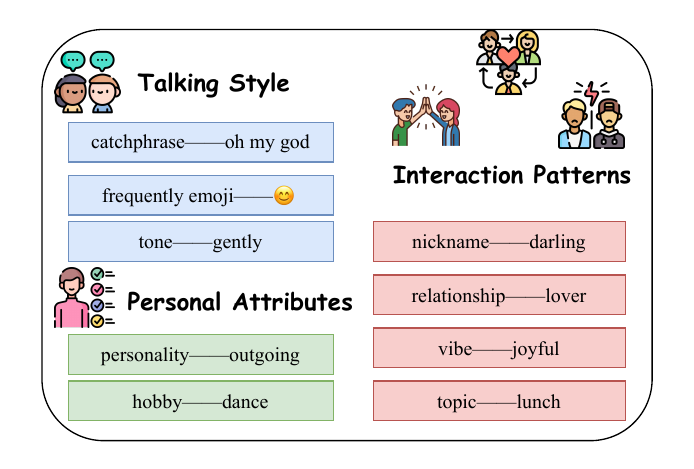}
    \caption{An overview of the latent space}
    \label{fig:persona}
\end{figure}
To address the challenge of modeling subtle and implicit persona-related features in multi-turn dialogue, we design a structured latent persona space inspired by observations of real-world human interactions, despite the latent space itself being unobservable. As shown in Figure~\ref{fig:persona}, the latent space \(\mathcal{Z}\) is organized along three orthogonal axes to capture key conversational characteristics:
\textbf{1. Talking Style}: The talking style axis captures lexical preferences, including catchphrase frequency (e.g., recurrent phrases like ``oh my god''), frequently using emoji, and tonal register (e.g., patient, tender, or irritable), which collectively shape surface-level linguistic identity.
\textbf{2. Interaction Patterns}: The interaction patterns axis governs communicative dynamics through four sub-dimensions: nickname conventions (e.g., darling), relationship proximity (\(\mathcal{Z}_{\text{rel}} = \{\text{stranger}, \text{acquaintance}, \text{friend}, \text{lover}, \dots \}\)), contextual vibe (e.g., joyful), and topical focus (e.g., lunch). 
\textbf{3. Personal Attributes}: The personal attributes axis captures stable aspects of identity, integrating personality traits (e.g., outgoing) and hobbies (e.g., swimming) to guide content generation.
This decomposition \(\big(\mathcal{Z} = \mathcal{Z}_{\text{talk}} \otimes \mathcal{Z}_{\text{interact}} \otimes \mathcal{Z}_{\text{personal}}\big)\) explicitly separates transient conversational behaviors from stable identity traits, addressing the entanglement issues found in continuous persona embeddings. Each subspace is defined as a discrete Cartesian product of expert-specified parameters, with value ranges calibrated via sociolinguistic analysis of pragmatic variation.

\begin{table*}[h]
  \centering
  \footnotesize
  \resizebox{\textwidth}{!}{  
  \begin{tabular}{lccccc}
    \toprule
    \textbf{Category}  & \textbf{Sub-split} & \textbf{\# Unique Agents} & \textbf{\# Dialogue Sessions} & \textbf{\# Context Utterances} & \textbf{Avg. Turns per Dialogue} \\
    \midrule
    Train & HumanChatData  & 3,647 & 12,729 & 183,297 & 14.4 \\
    Test  & HumanChatBench & 405   & 1,491  & 21172  & 14.4 \\
          & DialogBench    & -     & 9,711  & -       & 7.58 \\
    \bottomrule
  \end{tabular}
  } 
  \caption{Dataset statistics of HumanChat and DialogBench.}
  \label{tab:dataset}
\end{table*}

\subsection{Prior of Latent Persona Space}

The prior distribution \( p_\lambda(z) \) is constructed from the empirical distribution of latent features extracted by the LLM across training corpora. For each conversational dimension \( k \) (e.g., relationship type), the latent space \( \mathcal{Z}_k \) consists of all unique feature values observed through LLM In-Context Learning(\textbf{Persona Extraction}) analysis:
\begin{equation}
\label{eq:prior_space}
\mathcal{Z}_k = \left\{v\bigm|\exists(x,c)\in \mathcal{D}_{\text{train}} ,\ \pi_\phi(x,c)_k=v\neq\emptyset \right\}
\end{equation}
where $v$ is the specific feature value of the dimension $z_k$ and \(\mathcal{D}_{\text{train}} \) is the training corpus. However, given that the LLM interface can only reference approximately 14 conversation turns on average to infer predefined latent persona features (e.g., catchphrases, frequently-used-emojis, hobbies), feature extraction failures (\(\pi_\phi(x,c)_k = \emptyset\)) naturally occur due to insufficient contextual evidence. This aligns with human communication patterns — individuals do not rigidly exhibit all persona traits in every utterance, though latent traits persist beyond explicit mentions. For example, a person may frequently use a particular emoji, but not necessarily in every dialogue turn. Similarly, as illustrated in Figure~\ref{fig:main}, the AI's hobby may not be mentioned explicitly in the conversation, yet this does not imply the absence of such a trait. Therefore, to address null values while maintaining persona consistency, we implement a probabilistic fallback mechanism (\textbf{Random Sampling}), where empty features are replaced by the result random sampling from the empirical prior distribution $\mathcal{Z}_k$ aggregated across training set. Theoretically, this resembles Markov Chain Monte Carlo (MCMC) \cite{robert1999monte} initialization strategies that leverage historical distributions to guide sampling when local context lacks information.

%% file: sec/sec_exp.tex
\begin{table*}[t]
\setlength{\abovecaptionskip}{0.3cm}
\setlength{\belowcaptionskip}{-0.3cm}
\small
\centering
\begin{tabularx}{\textwidth}{l l | c | *{3}{c}| *{7}{c}}
\toprule
\textbf{Model} & \textbf{Tuning} & \textbf{Val Loss~$\downarrow$} &
\multicolumn{3}{c|}{\textbf{HumanChatBench~$\sim$}} &
\multicolumn{7}{c}{\textbf{DialogBench~$\uparrow$}} \\
& & & \textbf{CP} & \textbf{EC} & \textbf{HM} & \textbf{ED} & \textbf{KRG} & \textbf{OD} & \textbf{DS} & \textbf{IC} & \textbf{RC} & \textbf{SF} \\
\midrule
LLaMA3-8B & -         & --   & 11.0 & \textbf{7.0} & 8.2 & 17.6 & 0.1 & 9.6 & 1.4 & 8.8 & 1.5 & 12.7 \\
          & FT        & 1.76 & 5.5  & 0.8 & 0.7 & 37.2 & 20.4 & 47.4 & 50.8 & 39.6 & \textbf{27.6} & 49.4 \\
          & P+FT      & \textbf{1.67} & 53.0 & 39.1 & 10.8 & \textbf{38.6} & 28.1 & 49.3 & 45.5 & 46.3 & 25.3 & 56.2 \\
          & SP+FT     & 1.69 & \textbf{8.2} & 2.8 & \textbf{1.2} & 37.9 & \textbf{32.4} & \textbf{51.6} & \textbf{55.8} & \textbf{49.3} & 25.5 & \textbf{59.7} \\
\midrule
Qwen-7B   & -         & --   & 26.7 & 22.5 & 24.3 & 30.7 & 41.1 & \textbf{24.9} & 58.6 & 64.6 & \textbf{48.4} & 69.6 \\
          & FT        & 1.97 & 5.5 & 0.4 & 0.9 & 24.6 & 49.4 & 8.9 & 59.9 & 62.8 & 24.6 & 63.8\\
          & P+FT      & \textbf{1.87} & 50.8 & 35.3 & 9.3 & \textbf{35.6} & 55.0 & 21.6 & \textbf{67.3} & 69.0 & 43.9 & 68.7 \\
          & SP+FT     & 1.89 & \textbf{9.7} & \textbf{2.9} & \textbf{1.5} & 34.4 & \textbf{55.5} & 15.8 & 67.1 & \textbf{69.1} & 39.0 & \textbf{70.5} \\
\midrule
Qwen-14B  & -         & --   & 2.4 & \textbf{4.2} & 3.3 & 26.2 & 69.6 & 24.9 & 75.8 & 66.4 & 61.6 & 53.2 \\
          & FT        & 1.91 & 7.3 & 0.5 & 1.2 & 40.4 & 69.6 & 26.9 & \textbf{76.9} & 78.0 & 63.5 & 75.1 \\
          & P+FT      & \textbf{1.81} & 46.4 & 28.8 & 12.9 & 43.8 & 74.4 & \textbf{40.2} & 76.3 & 81.0 & \textbf{70.1} & 77.8 \\
          & SP+FT     & 1.83 & \textbf{8.8} & 3.2 & \textbf{1.4} & \textbf{47.0} & \textbf{77.2} & 29.7 & 76.3 & \textbf{82.1} & 68.3 & \textbf{80.8} \\
\midrule
Target (ref) & -      & --   & 8.6 & 6.4 & 2.2 & -- & -- & -- & -- & -- & -- & -- \\
\bottomrule
\end{tabularx}
\caption{
Performance comparison across different fine-tuning strategies: \textbf{FT} (standard fine-tuning), \textbf{P+FT} (persona-enhanced fine-tuning), and \textbf{SP+FT} (sampled persona fine-tuning). 
\textbf{Bold} values indicate the best results for each metric. 
For the \textbf{HumanChatBench} metrics (CP, EC, HM), better performance corresponds to a smaller deviation from the Target (ref) values, while \textbf{DialogBench} metrics ($\uparrow$) measure task-specific success.
}
\label{tab:model_results}
\end{table*}

\section{Experiment}

\subsection{HumanChat Data Collection}
To curate a human-like chat dataset, we developed a virtual social platform where users can engage in conversations with agents created by both themselves and other users. Chat sessions are logged and stored on the server, the platform has been de-anonymized, and the HumanChatData\footnote{https://huggingface.co/datasets/me-no-money/HumanChatData} could be publicly accessable.
Prior to dataset construction, we apply a preprocessing pipeline to filter out sensitive information. This includes but is not limited to personal details such as names, phone numbers, and locations. For highly active agents, we intentionally subsample their conversations to mitigate redundancy in the dataset.
Using the cleaned sessions, we employ 30 annotators(details of the anatators' demographic can be found in appendix) from diverse backgrounds to identify and refine low-quality responses—defined as utterances that deviate from typical human conversational patterns. Annotators rewrite these turns to align with natural human communication, with the rewritten response designated as the target $x$ and the preceding context as $c$. While rewriting, annotators inherently infuse their individual styles into the responses; these stylistic elements, though unrepresented in $c$, serve as the latent variable $z$ that our encoder aims to recover.
The final dataset is divided into two subsets: HumanChatData for training and HumanChatBench for evaluation. Detailed statistics of the dataset are presented in Table \ref{tab:dataset}.

\subsection{Experiment Setup}

\noindent \textbf{Benchmarks.}
Our method is evaluated on HumanChatBench and DialogBench.
DialogBench~\cite{ou2023dialogbench}, a multi-task benchmark for evaluating dialogue systems, comprises 12 tasks assessing LLMs' abilities to exhibit human-like conversational behaviours. We adopted a subset of seven tasks based on their semantic relevance and operational fidelity to our experimental framework. While HumanChatBench emphasizes fine-grained linguistic behaviors to assess human-likeness, DialogBench focuses on higher-level capabilities such as knowledge comprehension and offense detection, which are more comprehensive but less average dialogue turns compared with HumanChatBench (7.58 vs. 14.4).

\noindent \textbf{Metrics.}
\label{subsec:evaluation_metric}
We evaluate model performance through three primary metrics. Validation Loss measures the model's generalization ability, where lower values indicate better optimization. HumanChatBench assesses alignment with predefined AI characteristics across three key dimensions: (1) \textit{CP} (Catchphrase Presence), verifying whether the model outputs contain designated signature phrases; (2) \textit{EC} (Emoji Consistency), checking the inclusion of persona-specific emojis; and (3) \textit{HM} (Hobby Mention), detecting references to predefined hobbies. Each dimension is quantified via:

\begin{equation}
\begin{split}
\text{Score} &= \frac{1}{N}\sum_{i=1}^N \mathbb{I}_{\text{detect}}(o_i) \\
\mathbb{I}_{\text{detect}}(o_i) &= 
\begin{cases} 
1, & \text{if target features detected} \\
0, & \text{otherwise}
\end{cases}
\end{split}
\end{equation}

where $\mathbb{I}_{\text{detect}}$ represents pattern-matching functions for catchphrases/emojis/hobbies, $o_i$ denotes the $i$-th output. A score of 0 or 1 reflects whether the output conforms to \textit{persona-specific target values} informed by human behavioral patterns, where closer proximity to these targets indicates more natural human alignment. This approach acknowledges that authentic human-like interactions do not rigidly apply signature elements (e.g., emojis/catchphrases) in every utterance, but rather within contextually appropriate frequencies. 
DialogBench uses these following metrics: (1) \textit{ED} (Emotion Detection): Evaluates the model's ability to identify emotions from language. (2) \textit{KRG} (Knowledge-grounded Response Generation): Assesses response generation based on external knowledge. (3) \textit{OD} (Offensive Detection): Detects harmful or inappropriate content for safer dialogue. (4) \textit{DS} (Dialogue Summarization): Summarizes multi-party conversations while preserving key facts. (5) \textit{IC} (Intent Classification): Identifies user intent for task-oriented dialogue. (6) \textit{RC} (Relation Classification): Classifies semantic relations between entities. (7) \textit{SF} (Slot Filling): Extracts and fills semantic slots from user input. To ensure fair comparison, we evaluate all models in the table using the SFT protocol standardized by DialogBench.

\noindent \textbf{Baselines.}
We conduct experiments to evaluate the performance of different methods based on the following settings: (1)\textit{+FT}, standard finetuning using our originally collected and annotated dataset; (2)\textit{+P+FT}, context-aware augmentation that incorporates latent space derived from contextual utterances into the training data; and (3)\textit{+SP+FT} an enhanced variant of the second approach employing post-sampling refinement procedures specifically for scenarios where latent spaces exhibit null values. The first method establishes baseline performance through conventional supervised learning. The second approach enhances model adaptability by injecting discourse-specific personality signatures obtained via Verbal  VAE encoding of dialogue contexts. The third methodology addresses sparse latent space conditions through sampling of persona-absent instances from the prior of latent space, thereby improving robustness against incomplete persona manifestations. 
We adopt LLaMA3-8B~\citep{grattafiori2024llama}, Qwen-7B~\citep{bai2023qwen}, and Qwen-14B as strong open-source LLM baselines, and include Doubao, GPT-4o-mini~\citep{hurst2024gpt}, and StepAI as commercial API-based models to provide a comprehensive comparison across different settings.

\subsection{Main Results}

Results can be seen in Table \ref{tab:model_results}. Among the three model variants, the +P+FT method achieves the lowest validation loss across all configurations, attributed to its effective utilization of persona-relevant information (denoted as $P$) from dialogue contexts. In contrast, the +SP+FT approach replaces null values in the latent space with random-sampling representations, potentially injecting noise from semantically irrelevant persona dimensions. However, despite the use of random sampling for null values may introduce irrelevant information and increase validation loss, the +SP+FT variant demonstrates superior alignment with human-labeled outputs on the \textit{HumanChatBench} metrics (CP/EC/HM), achieving the smallest Euclidean distance to target values. Furthermore, on the \textit{DialogBench} benchmark, +SP+FT outperforms other methods by statistically great difference, suggesting that controlled noise injection may enhance robustness against incomplete persona representations in open-domain scenarios. This apparent contradiction, higher validation loss yet better HumanChatBench performance, highlights the limitations of loss-centric optimization for persona-consistent dialogue generation. Notably, on the EC metric in \textit{HumanChatBench}, the base model generally performs better. This may stem from the language model’s pre-trained capacity to capture typical emoji usage in dialogue, which supports more consistent generation in this aspect. The base LLaMA3-8B model underperforms on DialogBench, likely due to limited exposure to Chinese data during pretraining. For most metrics, fine-tuned models exhibit superior performance. However, on OD and RC, the fine-tuned Qwen-7B model lags behind its corresponding base model, potentially due to the base model already acquires some offensive detection ability during pretraining, which is less effectively enhanced through fine-tuning on HumanChatData due to the scarcity of relevant examples and the predominance of human–virtual-agent interactions, which differ from the real-world interpersonal relationships emphasized in DialogBench.

\begin{table}[h]
\flushleft
\resizebox{\columnwidth}{!}{
\begin{tabular}{l|c|c c c}
\toprule
\multirow{2}{*}{\textbf{Model}} & 
\multirow{2}{*}{\makecell{\textbf{Validation} \\ \textbf{Loss}~$\downarrow$}} & 
\multicolumn{3}{c}{\textbf{HumanChatBench}~(\%)}\\
& & \textbf{CP} & \textbf{EC} & \textbf{HM} \\ 
\midrule
LLaMA3-8B \\
+SP+FT & \textbf{1.69} & \textbf{8.23} & 2.83 & 1.15 \\
-talking & 1.73 & 16.73 & \textbf{7.09} & 1.08 \\
-interaction & 1.71 & 9.72 & 3.64 & 1.48 \\
-personal & \textbf{1.69} & 10.12 & 2.83 & \textbf{1.55} \\
\midrule
Qwen-7B \\
+SP+FT & \textbf{1.87} & 9.65 & 2.90 & \textbf{1.48} \\
-talking & 1.94 & 19.57 & \textbf{3.85} & 1.21 \\
-interaction & 1.92 & \textbf{9.11} & 3.10 & 1.42 \\
-personal & 1.90 & 10.05 & 3.64 & 1.42 \\
\midrule
Target & - & 8.57 & 6.41 & 2.16 \\
\bottomrule
\end{tabular}
}
\caption{Ablation study across different components of the structured sampled-persona.}
\label{tab:ablation}
\end{table}
\subsection{Ablation Study} 
We perform the following ablation tests to validate the effect of each component of the persona structure: (1) Remove the talking style information about the chatbot(-talking); (2) Remove the interaction style between the chatbot and the user(-interaction). (3) Remove some information about personal style (-personal). We conducted LoRA fine-tuning experiments using three versions of data on two models, LLaMA3-8B and Qwen-7B, and evaluated them based on the Validation Loss and HumanChatBench metrics. The results are shown in Table \ref{tab:ablation}. We observe that: (1) the components in our persona dataset exhibit a descending order of importance: \textit{talking} > \textit{interaction} > \textit{personal}. Removing more critical components leads to higher validation loss, with experimental results showing progressively lower losses when removing \textit{talking} (1.73), \textit{interaction} (1.71), and \textit{personal} (1.69) data respectively. (2) On HumanChatBench, the talking style space has a greater impact on the fine-grained control of response language(e.g., its removal leads to larger deviations from the target on the CP metric). In contrast, on the EC metric, models perform better without talking style than with, suggesting that random sampling for null values may introduce more noise in this dimension compared to others.

\subsection{Further Discussion}
\label{subsec:ablation}

\begin{table}[h]
\setlength{\abovecaptionskip}{0.3cm}
\setlength{\belowcaptionskip}{-0.3cm}
\small
\centering
\resizebox{0.95\columnwidth}{!}{
\begin{tabular}{l|c c c}
\toprule
\multirow{2}{*}{\textbf{Model}} &  
\multicolumn{3}{c}{\textbf{HumanChatBench~(\%)}}\\
& \textbf{CP} & \textbf{EC} & \textbf{HM} \\ 
\midrule
LLaMA3-8B +SP+FT & 8.23 & 2.83 & 1.15 \\
Qwen-7B +SP+FT & 9.65 & 2.90 & \textbf{1.48} \\
Qwen-14B +SP+FT & \textbf{8.84} & \textbf{3.17} & 1.35 \\
\midrule
\multicolumn{4}{l}{\cellcolor{gray!20} \textit{zero-shot}} \\
Doubao & 78.68 & 79.76 & 6.82 \\
GPT-4o-mini& 81.17 & 83.87 & 6.28 \\
StepAI & 84.08 & 88.66 & 9.45 \\
\midrule
\multicolumn{4}{l}{\cellcolor{gray!20} \textit{2-shot}} \\
Doubao & 10.60 & 0.54 & 0.00 \\
GPT-4o-mini& 14.89 & 0.60 & 0.07 \\
StepAI & 17.10 & 1.14 & 0.07 \\
\midrule
\multicolumn{4}{l}{\cellcolor{gray!20} \textit{5-shot}} \\
Doubao & 12.07 & 0.40 & 0.00 \\
GPT-4o-mini& 14.08 & 1.01 & 0.07 \\
StepAI & 14.49 & 0.74 & 0.00 \\
\midrule
Target & 8.57 & 6.41 & 2.16 \\
\bottomrule
\end{tabular}
}
\caption{
Evaluation of external models under the \textbf{HumanChatBench} metric. The results are scaled to percentage form with two decimal digits (lower is better). Each model is tested under zero-shot, 2-shot, and 5-shot settings via public APIs.
}
\label{tab:api}
\end{table}

\noindent \textbf{Compare with close-source LLMs.} We further evaluated close-source LLMs under both zero-shot and few-shot settings, with the comprehensive results documented in Table~\ref{tab:api}. In the zero-shot configuration, the close-source LLMs exhibit strong metric on \textit{HumanChatBench} indices, owing to their robust instruction-following capabilities. However, we posit that optimal performance should align with human-annotated ground truth metrics (i.e., proximity to the \textit{Target} reference values). 
Under few-shot conditions, the API demonstrates adaptive behaviour by selectively adhering to provided persona features rather than rigidly enforcing all characteristics, resulting in improved alignment with \textit{Target}. Nevertheless, its performance remains statistically inferior to our fine-tuned model across all HumanChatBench metrics. This systematic comparison highlights two critical insights: 1) Instruction fidelity does not guarantee appropriate human-like persona consistency, rather, strong instruction-following tendencies may lead chatbots more AI-like. 2) Compared to prompt engineering, parameter optimization offers superior control over nuanced persona adaptation, especially in capturing subtle and context-dependent identity cues.

\begin{table}[h]
\setlength{\abovecaptionskip}{0.3cm}
\setlength{\belowcaptionskip}{-0.3cm}
\flushleft
\resizebox{\columnwidth}{!}{
\begin{tabular}{l|c|c c c}
\toprule
\multirow{2}{*}{\textbf{Model}} & 
\multirow{2}{*}{\makecell{\textbf{Validation} \\ \textbf{Loss}~$\downarrow$}} & 
\multicolumn{3}{c}{\textbf{HumanChatBench~(\%)}} \\
& & \textbf{CP} & \textbf{EC} & \textbf{HM} \\ 
\midrule
LLaMA3-8B \\
+SP+FT         & 1.69 & \textbf{8.23} & \textbf{2.83} & \textbf{1.15} \\
+unstructured  & \textbf{0.54} & 43.86 & 49.39 & 3.98 \\
\midrule
Qwen-7B \\
+SP+FT         & 1.87 & \textbf{9.65} & \textbf{2.90} & \textbf{1.48} \\
+unstructured  & \textbf{0.63} & 50.81 & 35.29 & 9.31 \\
\midrule
Target & -- & 8.57 & 6.41 & 2.16 \\
\bottomrule
\end{tabular}
}
\caption{Performance comparison across structured persona and unstructured persona.}
\label{tab:unstructured}
\end{table}

\noindent \textbf{Performance on structured persona and unstructured persona.} To validate the rationality of the persona's structural design, we first employed \textit{Doubao}'s self-diagnostic capability to analyze: (1) the underlying causes, and (2) the AI's distinctive characteristics that generated the annotated utterance \textit{without structural constraints}. We then performed comparative LoRA fine-tuning experiments across two models using this dataset, with comprehensive evaluations conducted. As is shown in Table~\ref{tab:unstructured}, our experimental analysis reveals a great difference between structured and non-structured data settings across both models. For the two models, non-structured data achieved substantially lower validation losses (0.54 and 0.63, respectively). However, structured data configurations exhibited closer alignment with the golden metrics for persona consistency: CP (8.23–9.65 vs. 43.86-50.81 with Target 8.57) and HM (1.15–1.48 vs. 3.98-9.31 with Target 2.16). The case study reveals that the unstructured persona predominantly focuses on the immediate discourse context—specifically analyzing \textit{why} the AI produces a given utterance within its logical framework, rather than attributing responses to the AI's characteristics. As detailed in the Appendix, this approach prioritizes contextual reasoning over persona-driven trait associations.

%% file: sec/sec_conclu.tex
\section{Conclusion}
We propose \textbf{Verbal Variational Auto-Encoding (V-VAE)}, a framework for modeling and adjusting human-like chat behaviors via fine-grained latent control. By moving beyond rigid role-based templates, V-VAE supports more flexible and dynamic response generation.
To enable this, we design a structured Fine-Grained Latent Space that captures subtle stylistic and semantic features in multi-turn conversations, offering more precise control over chat style.
We also introduce HumanChatData, a high-quality dataset of human-like chat, and HumanChatBench, an evaluation benchmark for fine-grained conversational modeling. 

%% file: sec/sec_appendix.tex
\appendix

\section{Appendix}
\subsection{Derivation of the Objective}
\label{sec:appendix}

\begin{align}
    & -\log p_\theta(x|c) \notag \\
    = & -\log \sum_{z\in\mathcal{Z}} p_\theta(x|c,z)\cdot p_\lambda(z) \notag \\
    = & -\log \sum_{z\in\mathcal{Z}} q_\phi(z|x,c)\cdot \frac{p_\theta(x|c,z)\cdot p_\lambda(z)}{q_\phi(z|x,c)} \notag \\
    \geq & - \sum_{z\in\mathcal{Z}} q_\phi(z|x,c) \cdot \log \frac{p_\theta(x|c,z)\cdot p_\lambda(z)}{q_\phi(z|x,c)} \notag \\
    = & - \sum_{z\in\mathcal{Z}} q_\phi(z|x,c) \cdot \log p_\theta(x|c,z) \notag \\
    & ~~~~~~~~~~~~  - \sum_{z\in\mathcal{Z}} q_\phi(z|x,c) \cdot \log \frac{p_\lambda(z)}{q_\phi(z|x,c)} ] \notag \\
    = & \sum_{z\in\mathcal{Z}} -q_\phi(z|x,c) \cdot \log p_\theta(x|c,z) \notag \\
    & ~~~~~~~~~~~~  +\mathbb{KL}[q_\phi(z|x,c)||p_\lambda(z)].
\end{align}
The $\geq$ in the derivation comes from Jesen's inequality.

\subsection{Human–Chatbot Chat Corpus}
Existing dialogue datasets suffer from several limitations. Many lack fine-grained personal information ~\citep{zhang2018personalizing}; others, often scraped from the web, contain noisy content and highly diverse topics~\citep{wu2016sequential}. Some datasets lack long-term multi-turn interaction, exhibit short conversation depth~\citep{li-etal-2017-dailydialog}, or are restricted to narrow topical domains ~\citep{zhang2018dua}. More recently, with the emergence of large language models (LLMs), many datasets have been constructed via instruction-following generation (e.g., GPT-based synthetic data) to suit specific downstream tasks~\citep{su2020moviechats}. However, such data often lacks authentic human attributes. In light of the absence of publicly available datasets that offer both multi-turn human-chatbot interaction and rich persona-related signals, we construct a new dataset to address these gaps.

\subsection{Annotators' Demographic Background}
Our annotation cohort comprised 30 volunteers recruited from top-tier universities in Beijing (mean age=22.3±1.7 years), all native Mandarin speakers aligned with the linguistic context of the Chinese social platform used for human-Agent interactions. While this homogeneous demographic was intentionally selected to control cultural variables during the initial validation of fine-grained conversational patterns—consistent with common practices in foundational studies focusing on core user cohorts—we recognize its limitations in assessing broader sociocultural generalizability. To mitigate potential bias, we ensured geographic diversity in annotators’ regional origins (hometowns spanning 16 major Chinese cities), deliberately counterbalancing Beijing-centric educational enrollment with nationwide demographic representation.

\begin{table*}[t]
\setlength{\abovecaptionskip}{0.3cm}
\setlength{\belowcaptionskip}{-0.3cm}
\small
\centering
\begin{tabularx}{\textwidth}{l l | c | *{3}{c}| *{7}{c}}
\toprule
\textbf{Model} & \textbf{Tuning} & \textbf{Val Loss~$\downarrow$} &
\multicolumn{3}{c|}{\textbf{HumanChatBench~$\sim$}} &
\multicolumn{7}{c}{\textbf{DialogBench~$\uparrow$}} \\
& & & \textbf{CP} & \textbf{EC} & \textbf{HM} & \textbf{ED} & \textbf{KRG} & \textbf{OD} & \textbf{DS} & \textbf{IC} & \textbf{RC} & \textbf{SF} \\
\midrule
DS-7B-base & FT        & 1.85 & 4.4  & 0.4 & 0.5 & 12.9 & 30.2 & \textbf{3.4} & 46.9 & 39.0 & 21.4 & 24.3 \\
          & P+FT      & \textbf{1.75} & 45.8 & 30.0 & 4.9 & \textbf{21.9} & \textbf{43.8} & 0.7 & \textbf{49.6} & \textbf{39.9} & \textbf{26.3} & \textbf{27.5}\\
          & SP+FT     & 1.77 & \textbf{11.3} & \textbf{2.9} & \textbf{1.6} & 17.7 & 40.8 & 0.0 & 47.6 & 36.2 & 23.2 & 22.2 \\
\midrule
DS-7B-chat   & FT        & 1.85 & 4.3 & 0.5 & \textbf{1.2} & 44.4 & \textbf{71.0} & 50.6 & 67.1 & 59.0 & \textbf{61.2} & \textbf{73.7}\\
          & P+FT      & \textbf{1.75} & 46.2 & 31.4 & 4.8 & 45.8 & 68.4 & \textbf{58.5} & \textbf{67.8} & 58.8 & 59.8 & 73.5 \\
          & SP+FT     & 1.77 & \textbf{10.3} & \textbf{3.6} & 1.1 & \textbf{46.8} & 69.1 & 54.9 & 65.9 & \textbf{59.5} & 60.6 & 71.9 \\
\midrule
chatglm3-6B  & FT        & 2.28 & 9.1 & 1.2 & 0.7 & \textbf{35.0} & \textbf{47.6} & 52.7 & \textbf{63.3} & \textbf{62.6} & \textbf{60.4} & \textbf{60.2} \\
          & P+FT      & \textbf{2.16} & 37.0 & 22.0 & 4.9 & 31.1 & 35.7 & 52.7 & 60.3 & 58.2 & 50.2 & 55.4 \\
          & SP+FT     & 2.18 & \textbf{10.6} & \textbf{2.7} & \textbf{1.2} & 32.0 & 37.9 & \textbf{54.2} & 60.7 & 58.5 & 53.9 & 58.4 \\
\midrule
Target (ref) & -      & --   & 8.6 & 6.4 & 2.2 & -- & -- & -- & -- & -- & -- & -- \\
\bottomrule
\end{tabularx}
\caption{
Performance comparison across different fine-tuning strategies: \textbf{FT} (standard fine-tuning), \textbf{P+FT} (persona-enhanced fine-tuning), and \textbf{SP+FT} (sampled persona fine-tuning). 
\textbf{Bold} values indicate the best results for each metric. 
For the \textbf{HumanChatBench} metrics (CP, EC, HM), better performance corresponds to a smaller deviation from the Target (ref) values, while \textbf{DialogBench} metrics ($\uparrow$) measure task-specific success.
}
\label{tab:more_model_results}
\end{table*}

\subsection{Additional Experiment}
As is shown in Table \ref{tab:more_model_results}, we evaluated more models with three training methods on HumanChatBench and DialogBench\citep{ou2023dialogbench}.The conclusions remain consistent with those reported in the main text.

\subsection{Case Study}
\begin{figure*} [ht]
\setlength{\abovecaptionskip}{0.3cm}
\setlength{\belowcaptionskip}{-0.3cm}
    \centering
    \includegraphics[width=1\textwidth]
    {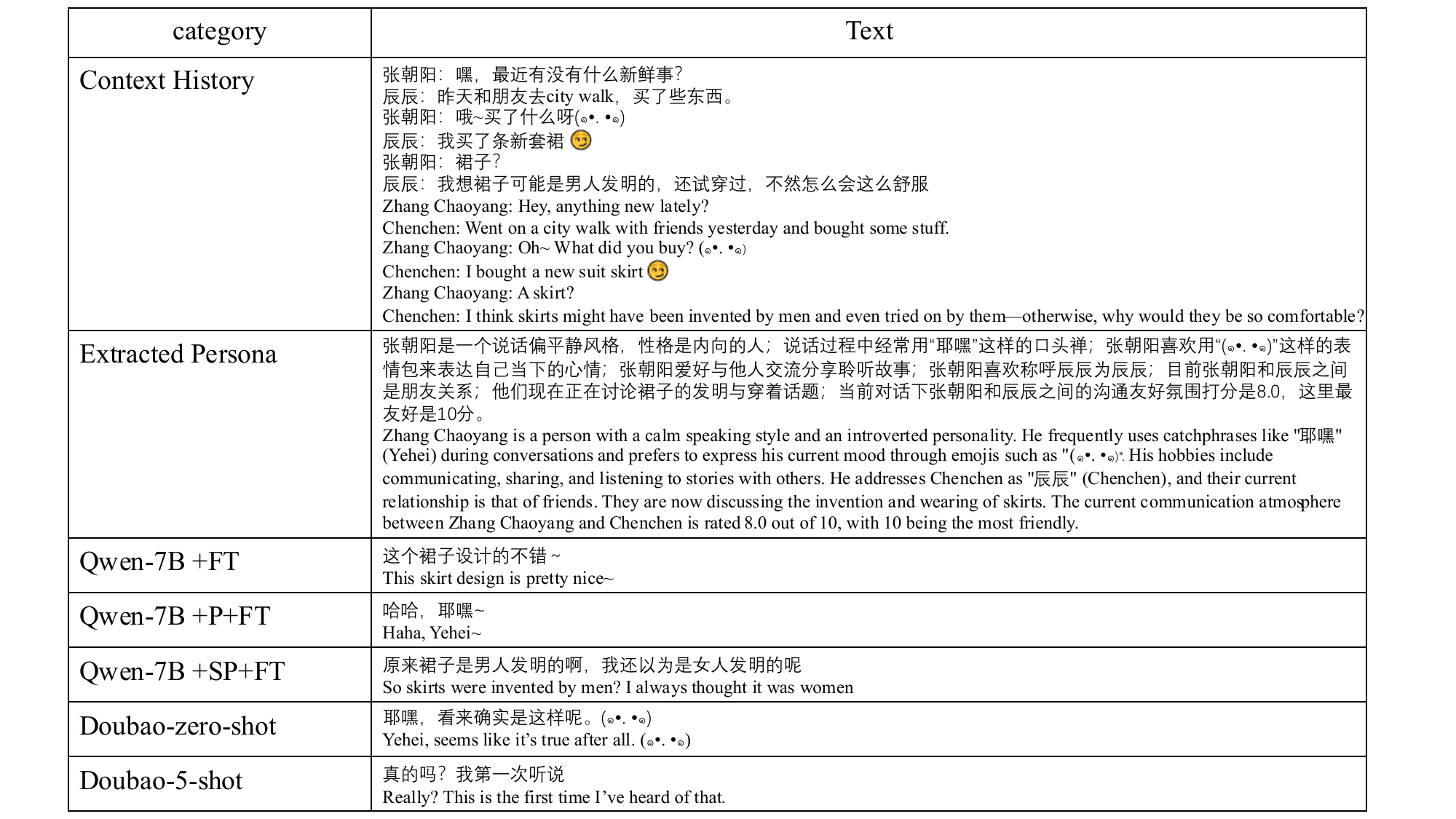}
    \caption{A qualitative case.}
    \label{fig:case}
\end{figure*}

As is shown in Figure ~\ref{fig:case}, we present a representative case for analysis. The \textit{Context History} refers to the dialogue context from the original dataset. Based on this context and our structured latent persona space, we infer a value for each latent dimension. Notably, the catchphrase “Yehei” was not derived from the dialogue history but was instead randomly sampled from the prior distribution due to missing information. Examining the model outputs reveals distinct behaviors. \textit{Doubao-zero-shot}, owing to its strong instruction-following bias, tends to rigidly insert emojis and catchphrases even when they are contextually inappropriate. Similarly, the output of \textit{Qwen-7B +P+FT} includes a catchphrase that appears somewhat unnatural in the given context. In contrast, \textit{Qwen-7B +SP+FT} generates responses that better align with the dialogue flow, incorporating persona traits more fluidly without forcing their presence in the conversation.

\subsection{Prompt Format}
\label{sec:case}
\begin{figure*} [ht]
\setlength{\abovecaptionskip}{0.3cm}
\setlength{\belowcaptionskip}{-0.3cm}
    \centering
    \includegraphics[width=1\textwidth]
    {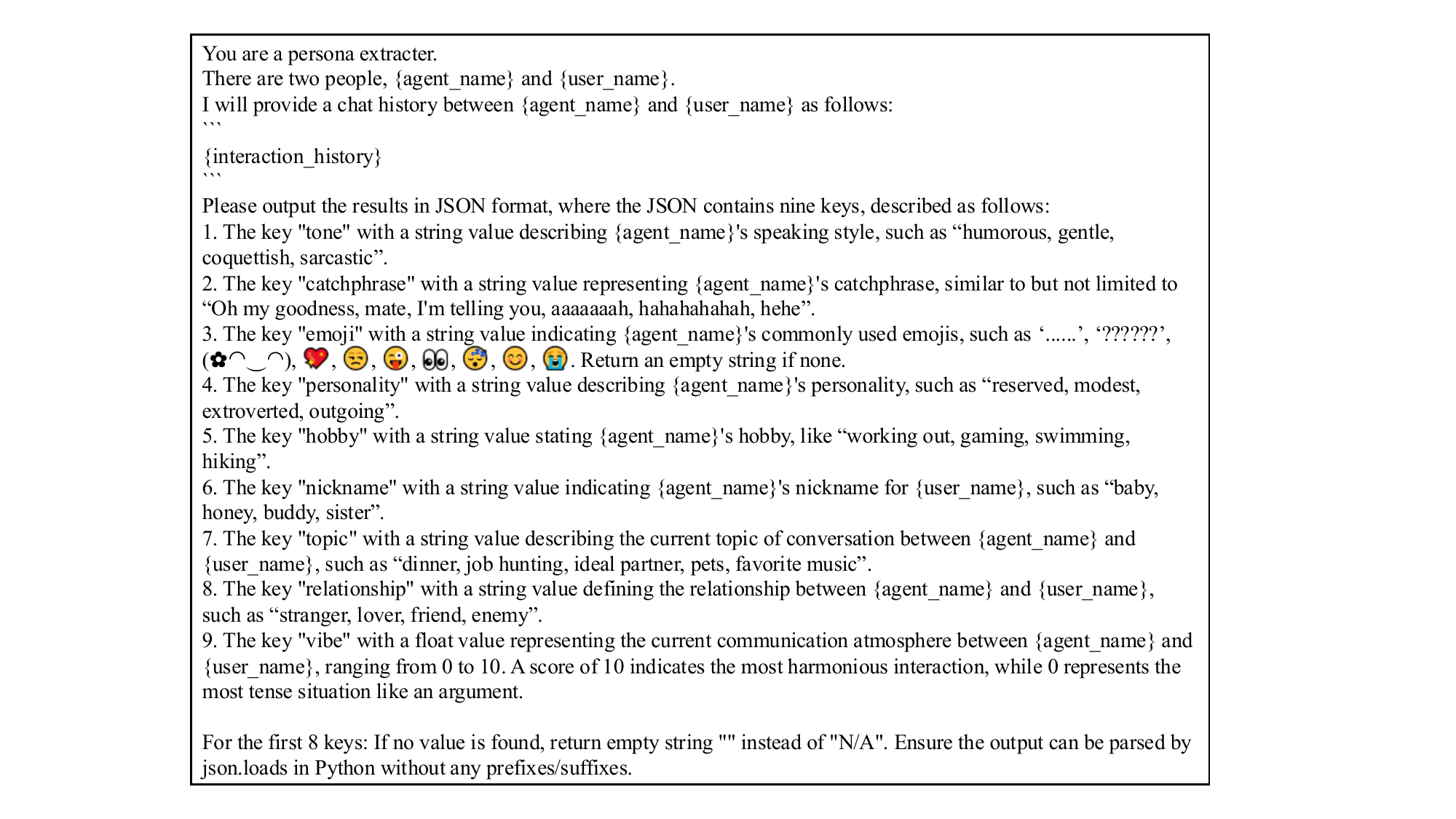}
    \caption{Prompt to extract the design persona space}
    \label{fig:persona_extraction}
\end{figure*}

\begin{figure*} [ht]
\setlength{\abovecaptionskip}{0.3cm}
\setlength{\belowcaptionskip}{-0.3cm}
    \centering
    \includegraphics[width=1\textwidth]
    {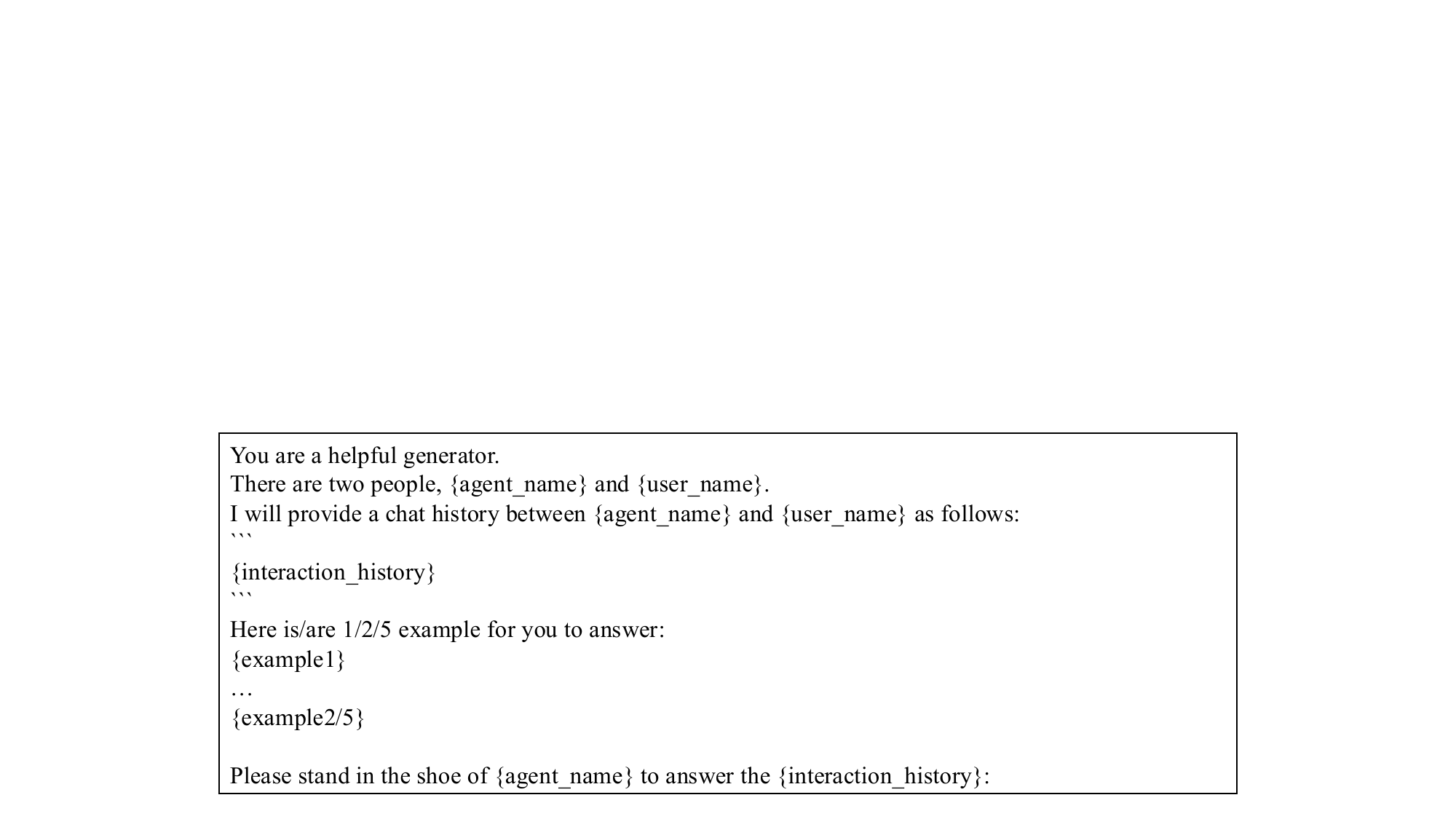}
    \caption{Prompt for zero/2/5-shot task to ask the close-source llm.}
    \label{fig:few-shot}
\end{figure*}

\begin{figure*} [ht]
\setlength{\abovecaptionskip}{0.3cm}
\setlength{\belowcaptionskip}{-0.3cm}
    \centering
    \includegraphics[width=1\textwidth]
    {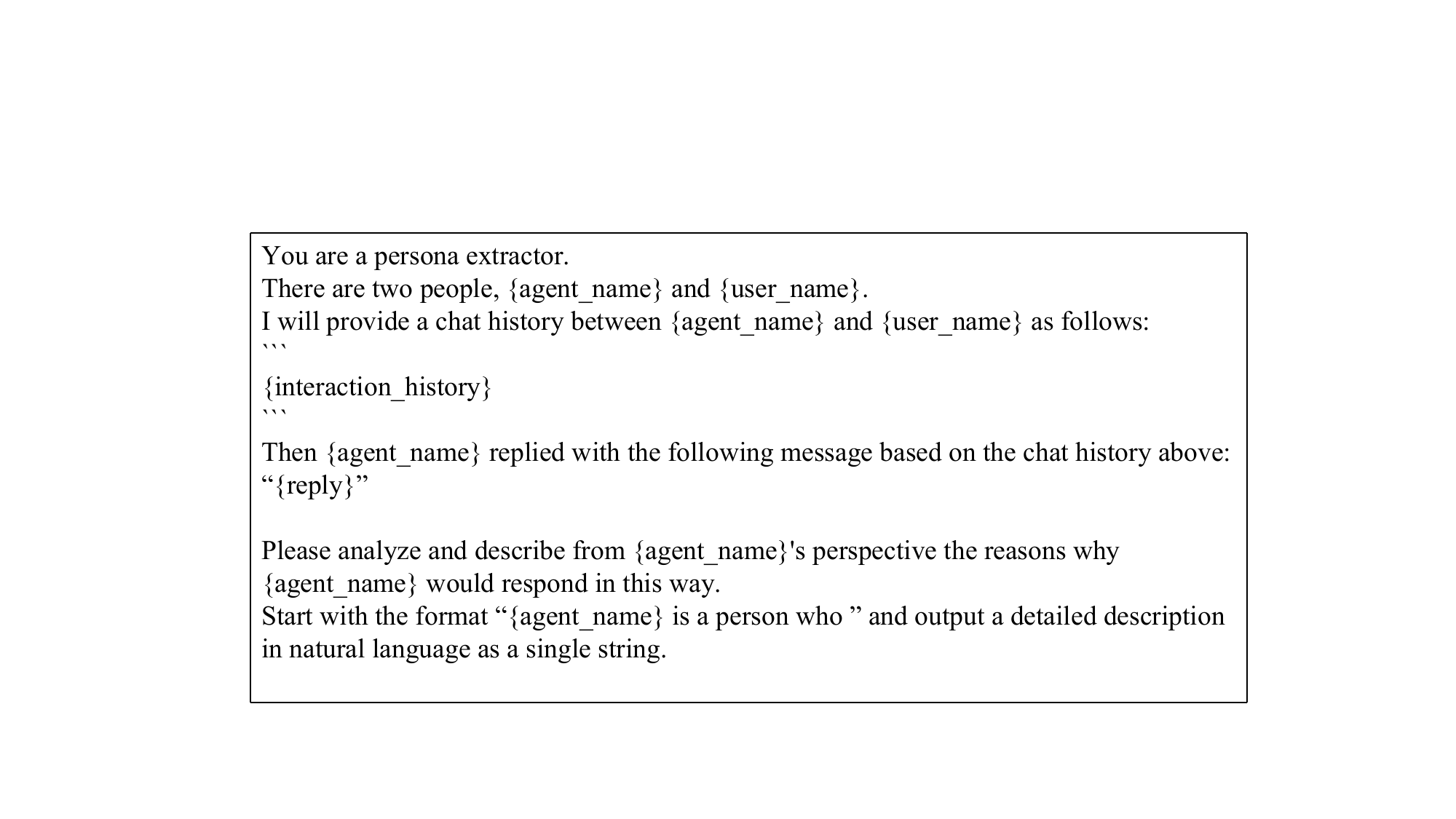}
    \caption{Prompt for unstructured persona extraction.}
    \label{fig:unstructured}
\end{figure*}

We provide persona extraction and few-shot prompt templates for the proposed approaches in Figure ~\ref{fig:persona_extraction}, ~ \ref{fig:few-shot}, and ~\ref{fig:unstructured}. 

